\documentclass[journal]{IEEEtran}
\ifCLASSINFOpdf
\else
\fi

\usepackage{times}
\usepackage{epsfig}
\usepackage{graphicx}
\usepackage{amsmath}
\usepackage{dsfont}
\usepackage{amssymb}
\usepackage{url}
\usepackage{amsfonts}
\usepackage{multirow}
\usepackage{makecell}
\usepackage{booktabs}
\usepackage{lineno}
\usepackage[table]{xcolor}
\usepackage{caption}
\usepackage{subcaption}

\begin{document}

\title{Multimodal Unified Attention Networks for Vision-and-Language Interactions}
\author{Zhou~Yu, \IEEEmembership{Member, IEEE},~
        Yuhao~Cui,~
        Jun~Yu, \IEEEmembership{Member, IEEE},~
        Dacheng~Tao, \IEEEmembership{Fellow, IEEE},~\\
        Qi~Tian \IEEEmembership{Fellow, IEEE}
\thanks{This work was supported in part by National Natural Science Foundation of China under Grant 61702143 and Grant 61836002, and in part by the Australian Research Council Projects under Grant FL-170100117. (Zhou Yu and Yuhao Cui contribute equally to this work. Jun Yu is the corresponding author.)}
\thanks{Z. Yu, Y. Cui and J. Yu are with Key Laboratory of Complex Systems Modeling and Simulation,
School of Computer Science and Technology, Hangzhou Dianzi University, P. R. China (e-mail: yuz@hdu.edu.cn; yujun@hdu.edu.cn; cuiyh@hdu.edu.cn}
\thanks{D. Tao is with the UBTECH Sydney AI Centre, School of Computer Science, Faculty of Engineering, The University of Sydney, Australia (e-mail: dacheng.tao@sydney.edu.au).}
\thanks{Q. Tian is with the Noah's Ark Lab, Huawei, P. R. China (e-mail: tian.qi1@huawei.com).}
        }
\markboth{Journal of \LaTeX\ Class Files,~Vol.~14, No.~8, August~2015}%
{Yu \MakeLowercase{\textit{et al.}}: Multimodal Unified Attention Networks for Vision-and-Language Interactions}
%



\maketitle


\begin{abstract}
Learning an effective attention mechanism for multimodal data is important in many vision-and-language tasks that require a synergic understanding of both the visual and textual contents. Existing state-of-the-art approaches use co-attention models to associate each visual object (\emph{e.g.}, image region) with each textual object (\emph{e.g.}, query word). Despite the success of these co-attention models, they only model inter-modal interactions while neglecting intra-modal interactions. Here we propose a general `unified attention' model that simultaneously captures the intra- and inter-modal interactions of multimodal features and outputs their corresponding attended representations. By stacking such unified attention blocks in depth, we obtain the deep Multimodal Unified Attention Network (MUAN), which can seamlessly be applied to the visual question answering (VQA) and visual grounding tasks. We evaluate our MUAN models on two VQA datasets and three visual grounding datasets, and the results show that MUAN achieves top level performance on both tasks without bells and whistles.
\end{abstract}


\section{Introduction}
Deep learning in computer vision and natural language processing has facilitated recent advances in artificial intelligence. Such advances drive research interest in multimodal learning tasks lying at the intersection of vision and language such as multimodal embedding learning \cite{zheng2019multimodal}\cite{yu2014discriminative}\cite{yang2018shared}, visual captioning \cite{xu2015show}\cite{song2017deterministic}, visual question answering (VQA) \cite{antol2015vqa} and visual grounding \cite{rohrbach2016grounding}, etc. In these tasks, learning a fine-grained semantic understanding of both visual and textual content is key to their performance.

The attention mechanism is a predominant focus of recent deep learning research. It aims to focus on certain data elements, and aggregate essential information to obtain a more discriminative local representation \cite{bahdanau2014neural,xu2015show}. This mechanism has improved the performance of a wide range of unimodal learning tasks (\emph{e.g.}, vision \cite{mnih2014recurrent,gregor2015draw,chen2016attention}, language \cite{luong2015effective,dozat2017deep,rush2015neural}) in conjunction with deep convolutional neural networks (CNNs) and recurrent neural networks (RNNs).

\begin{figure}
\begin{center}
\includegraphics[width=0.49\textwidth]{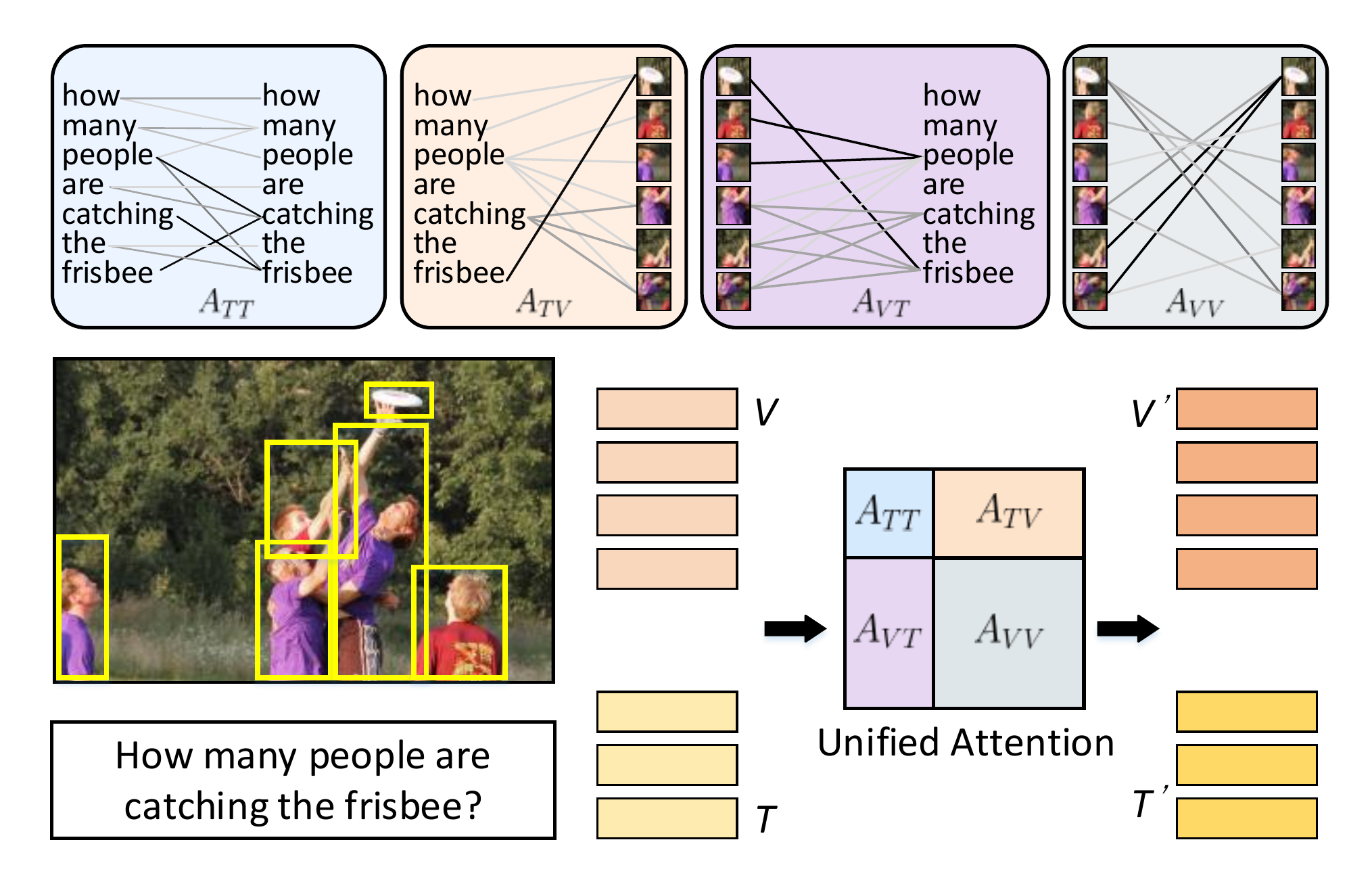}
\caption{Schematic of the proposed unified attention, which simultaneously models inter- and intra-modal interactions in a single framework. Given multimodal inputs $V$ and $T$, $A_{VV}$, $A_{TT}$ denote the intra-modal interactions within each modality, while $A_{VT}$ and $A_{TV}$ denote the inter-modal interactions across different modalities. $V'$ and $T'$ are the attended features for $V$ and $T$ respectively.}
\label{fig:unified_attention}
\end{center}
\end{figure}

For the multimodal learning tasks described above, attention learning considers the inputs from both the visual and textual modalities. Taking the VQA problem in Fig. \ref{fig:unified_attention} as an example, to correctly answer a question like \emph{`How many people are catching the frisbee'} for an image, the attention model should ideally learn to focus on particular image regions (\emph{i.e.}, the person near the frisbee). Such visual attention based models have become an integral component in many multimodal tasks that require fine-grained visual understanding \cite{xu2015show}\cite{yang2016stacked}\cite{fukui2016multimodal}. Beyond the visual attention models, recent studies have introduced co-attention models, which simultaneously learn the visual attention and textual attention to benefit from fine-grained representations for both modalities. Early approaches learned separate attention distributions for each modality in an iterative manner, neglecting the dense interaction between each question word and image region \cite{lu2016hierarchical}\cite{yu2017mfb}. To address this problem, dense co-attention models have been proposed to capture complete interactions between word-region pairs, which are further extended to form deep co-attention models \cite{kim2018bilinear}\cite{nguyen2018improved}.

Despite the success of the co-attention models in multimodal learning tasks, these models only consider inter-modal interactions (\emph{i.e.}, $A_{TV}$ or $A_{TV}$ in Fig. \ref{fig:unified_attention}) while neglecting intra-modal ones (\emph{i.e.}, $A_{TT}$ and $A_{VV}$). On the other hand, modeling intra-modal interactions has been proved to be beneficial for many unimodal learning tasks \cite{vaswani2017attention}\cite{devlin2018bert}\cite{wang2018non}\cite{hu2018relation}. We argue that intra-modal interactions within each modality provide complementary and important information to the inter-modal interactions.

Inspired by the famous self-attention model \cite{vaswani2017attention} in the NLP community, we naturally extend its idea for multimodal data and propose a \emph{unified attention} accordingly. Our unified attention model characterizes the intra- and inter-modal interactions jointly in a unified framework which we call the unified attention (UA) block (see Fig. \ref{fig:unified_attention}). The attention map learned from the UA block includes four relationships: the inter-modal interactions ($A_{VT}$ and $A_{TV}$) to build co-attention across different modalities, and the intra-modal interactions ($A_{VV}$ and $A_{TT}$) to build self-attention within each modality. The learned unified attention is further used to obtain the attended output features for multimodal inputs. By stacking such UA block in depth, we obtain the Multimodal Unified Attention Network (MUAN), which can be trained in an end-to-end manner to perform deep multimodal reasoning.

To evaluate the effectiveness of our proposed MUAN model, we apply it to for VQA and visual grounding. The quantitative and qualitative results on two VQA dataset VQA-v2 \cite{goyal2016making} and CLEVR \cite{johnson2017clevr}, and three visual grounding datasets RefCOCO \cite{kazemzadeh2014referitgame}, RefCOCO+ \cite{kazemzadeh2014referitgame} and RefCOCOg \cite{mao2016generation} show that MUAN achieves top level performance on both tasks without using any dataset specific model tuning.

In summary, we have made the following contributions in this study:
\begin{itemize}
  \item We extend the self-attention model for single modality to a unified attention model, which can characterize intra- and inter-modal interactions of multimodal data. By stacking such unified attention model (\emph{i.e.}, UA block) in depth, we obtain a neat multimodal unified attention network (MUAN), which can perform accurate multimodal reasoning.
  \item We modify the original self-attention model to a gated self-attention (GSA) model as the basic component for the UA block, which facilities more accurate and robust attention learning and leads to more discriminative features for specific tasks.
  \item We apply MUAN to two multimodal learning tasks, namely VQA and visual grounding. The results on five benchmark datasets show the superiority of MUAN over existing state-of-the-art approaches.
\end{itemize}

\section{Related Work}
We briefly review existing studies on VQA and visual grounding, and establish a connection between these two tasks by attention learning.

\noindent \textbf{Visual Question Answering (VQA).} VQA aims to answer a question in natural language with respect to a given image, so requires multimodal reasoning over multimodal inputs. Since Antol \emph{et al.} presented a large-scale VQA benchmark dataset with free-form questions \cite{antol2015vqa}, multimodal fusion and attention learning have become two major research focuses for VQA. For multimodal fusion, early methods used simple concatenation or element-wise multiplication between multimodal features \cite{zhou2015simple}\cite{antol2015vqa}. Fukui \emph{et al.} \cite{fukui2016multimodal}, Kim \emph{et al.} \cite{kim2016hadamard}, Yu \emph{et al.} \cite{yu2017mfb} and Ben \emph{et al.} \cite{ben2017mutan} proposed different approximated bilinear pooling methods to effectively integrate the multimodal features with second-order feature interactions. For attention learning, question-guided visual attention on image regions has become the de-facto component in many VQA approaches \cite{yang2016stacked}\cite{chen2015abc}. Chen \emph{et al.} proposed a question-guided attention map that projects the question embeddings to the visual space and formulates a configurable convolutional kernel to search the image attention region \cite{chen2015abc}. Yang \emph{et al.} proposed a stacked attention network to learn the attention iteratively \cite{yang2016stacked}. Some approaches introduce off-the-shelf object detectors \cite{ilievski2016focused} or object proposals \cite{shih2016look} as the candidates of the attention regions and then use the question to identify the relevant ones. Taken further, co-attention models that consider both textual and visual attentions have been proposed \cite{lu2016hierarchical}\cite{yu2017mfb}. Lu \emph{et al.} proposed a co-attention learning framework to alternately learn the image attention and question attention \cite{lu2016hierarchical}. Yu \emph{et al.} reduced the co-attention method into two steps, self-attention for a question embedding and the question-conditioned attention for a visual embedding \cite{yu2018beyond}. The learned co-attentions by these approaches are coarse, in that they neglect the interaction between question words and image regions. To address this issue, Nguyen \emph{et al.} \cite{nguyen2018improved} and Kim \emph{et al.} \cite{kim2018bilinear} introduced dense co-attention models that established the complete interaction between each question word and each image region.

\noindent \textbf{Visual Grounding.} Visual grounding (\emph{a.k.a.}, referring expression comprehension) aims to localize an object in an image referred to in query text. Most previous approaches follow a two-stage pipeline \cite{rohrbach2016grounding}\cite{yu2017joint}\cite{fukui2016multimodal}: 1) use an off-the-shelf object detector, such as Edgebox \cite{zitnick2014edge} or Faster R-CNN \cite{he2015deep} to generate a set of region proposals along with the proposal features for the input image; and 2) compute a matching score between each proposal feature and query feature and adopt the proposal (or its refined bounding box \cite{yu2018rethinking}) with the highest score as the referent. From the attention learning point of view, visual grounding represents a task of learning query-guided attention on the image region proposals. The aforementioned two-stage approaches are analogous to the visual attention models in VQA. Yu \emph{et al.} \cite{yu2018mattnet}, Zhang \emph{et al.} \cite{zhuang2018parallel} and Deng \emph{et al.} \cite{deng2018visual} also modeled the attention on question words along with visual attention, providing a connection to the co-attention model in VQA.

\noindent \textbf{Joint Modeling of Self- and Co-Attention.} Although extensive studies on self-attention and co-attention have been made by existing multimodal learning methods, the two kinds of attentions are usually considered solely. To the best of our knowledge, only a few attempts have modeled intra- and inter-modal interactions jointly. Li \emph{et al.} introduced a videoQA approach which used self-attention to learn intra-modal interactions of video and question modalities respectively, and then fed them through a co-attention block to model inter-modal interactions \cite{li2019beyond}. Gao \emph{et al.} presented a dynamic fusion framework for VQA with modeling intra- and inter-modal attention blocks. \cite{peng2018dynamic}. Yu \emph{et al} applied a modular co-attention network for VQA which stacked multiple self-attention and guided-attention blocks in depth to perform deep visual reasoning. In summary, all these methods models the self-attention and co-attention in two sequential stages, which is sub-optimal and may result in serious information lose. This inspires us to design a general unified attention framework to simultaneously model the two attentions in one stage.

\section{Multimodal Unified Attention}
In this section, we introduce the multimodal \emph{unified attention}, which is the basic component of our Multimodal Unified Attention Network (MUAN). Taking the multimodal input features $X$ from the image modality and $Y$ from the text modality, the unified attention outputs their corresponding attended features. In contrast to existing visual attention methods, which model unidirectional inter-modal interactions (\emph{i.e.}, $X\rightarrow Y$) \cite{fukui2016multimodal}\cite{kim2016hadamard}, or the co-attention methods, which model bidirectional inter-modal interactions (\emph{i.e.}, $X\leftrightarrow Y$) \cite{kim2018bilinear}\cite{nguyen2018improved}, our unified attention models the intra-modal and inter-modal interactions simultaneously (\emph{i.e.}, $X\rightarrow X$, $Y\rightarrow Y$ and $X\leftrightarrow Y$) in a general framework.

Inspired by the \emph{self-attention} model which has achieved remarkable performance in natural language processing \cite{vaswani2017attention}\cite{radford2018improving}\cite{devlin2018bert}, we design a unified attention model for multimodal data. Furthermore, to obtain more accurate attention map in the unified attention learning, we introduce a bilinear pooling based gating model to reweight the importance of input features, which can to some extent eliminate the irrelevant or noisy features.

\subsection{Gated Self-Attention}
The self-attention model proposed in \cite{vaswani2017attention} takes a group of input features $X=[x_1;...;x_m]\in\mathbb{R}^{m\times d_x}$ and outputs a group of attended features $F=[f_1,...,f_m]\in\mathbb{R}^{m\times d}$, where $m$ is the number of samples, $d_x$ and $d$ are the dimensionalities of input and output features, respectively. To achieve this goal, $X$ is first fed into three independent fully-connected layers.
\begin{equation}\label{eq:qkv}
Q = \mathrm{FC}_q(X),~K = \mathrm{FC}_k(X),~V = \mathrm{FC}_v(X) \\
\end{equation}
where $Q, K, V\in\mathbb{R}^{m \times d}$ are three feature matrices of the same shape, corresponding to the queries, keys, and values, respectively.

Given a query $q\in Q$ and all keys $K$, we calculate the dot-products of $q$ with $K$, divide each by a scaling factor $\sqrt{d}$ and apply the softmax function to obtain the attention weights on the values. In practice, the attention function can be computed on all queries $Q$ simultaneously, and in doing so we obtain the output features $F$ as follows:
\begin{align}
 \label{eq:sdp} A &= \mathrm{softmax}(\frac{QK^T}{\sqrt{d}})\\
 \label{eq:sa} F &= AV
\end{align}
where $A\in\mathbb{R}^{m \times m}$ is the attention map containing the attention weights for all query-key pairs, and the output features $F$ are the weighted summation of the values $V$ determined by $A$.

Learning an accurate attention map $A$ is crucial for self-attention learning. The scaled dot-product attention in Eq.(\ref{eq:sdp}) models the relationship between feature pairs. However, the importance of each individual features is not explicitly considered during attention learning. Consequently, irrelevant or noisy features may have a negative impact on the attention map, resulting in inaccurate output features. To address this problem, we introduce a novel \emph{gating model} into Eq.(\ref{eq:sdp}) to improve the quality of the learned attention. Inspired by the bilinear pooling models which have been in fine-grained visual recognition \cite{li2016factorized} and multi-modal fusion \cite{kim2016hadamard}, we design a gating model based on low-rank bilinear pooling to reweight the features of $Q$ and $K$ before their scaled dot-products:
\begin{equation}\label{eq:bilinear}
M = \sigma\left(\mathrm{FC}^g\left(\mathrm{FC}_q^g(Q)\odot \mathrm{FC}^g_k(K)\right)\right)
\end{equation}
where $\mathrm{FC}_q^g, \mathrm{FC}_k^g\in\mathbb{R}^{d \times d_g}$, $\mathrm{FC}^g\in\mathbb{R}^{d_g\times 2}$ are three independent fully-connected layers, and $d_g$ is the dimensionality of the projected space. $\odot$ denotes the element-wise product function and $\sigma(\cdot)$ the sigmoid function. $M\in\mathbb{R}^{m \times 2}$ corresponds to the two masks $M_q\in\mathbb{R}^m$ and $M_k\in\mathbb{R}^m$ for the features $Q$ and $V$, respectively.

The learned two masks $M_q$ and $M_k$ are tiled to $\tilde{M}_q, \tilde{M}_k\in\mathbb{R}^{m\times d}$ and then used to formulate a gated self-attention (GSA) model as follows:
\begin{align}
\label{eq:gdp} &A^g = \mathrm{softmax}(\frac{(Q\odot \tilde{M}_q)(K\odot \tilde{M}_k)^T}{\sqrt{d}}) \\
\label{eq:gsa} &F = A^gV
\end{align}

\captionsetup[subfigure]{font=small}
\begin{figure}
    \centering
    \begin{subfigure}[h]{0.66\columnwidth}
        \includegraphics[width=\linewidth]{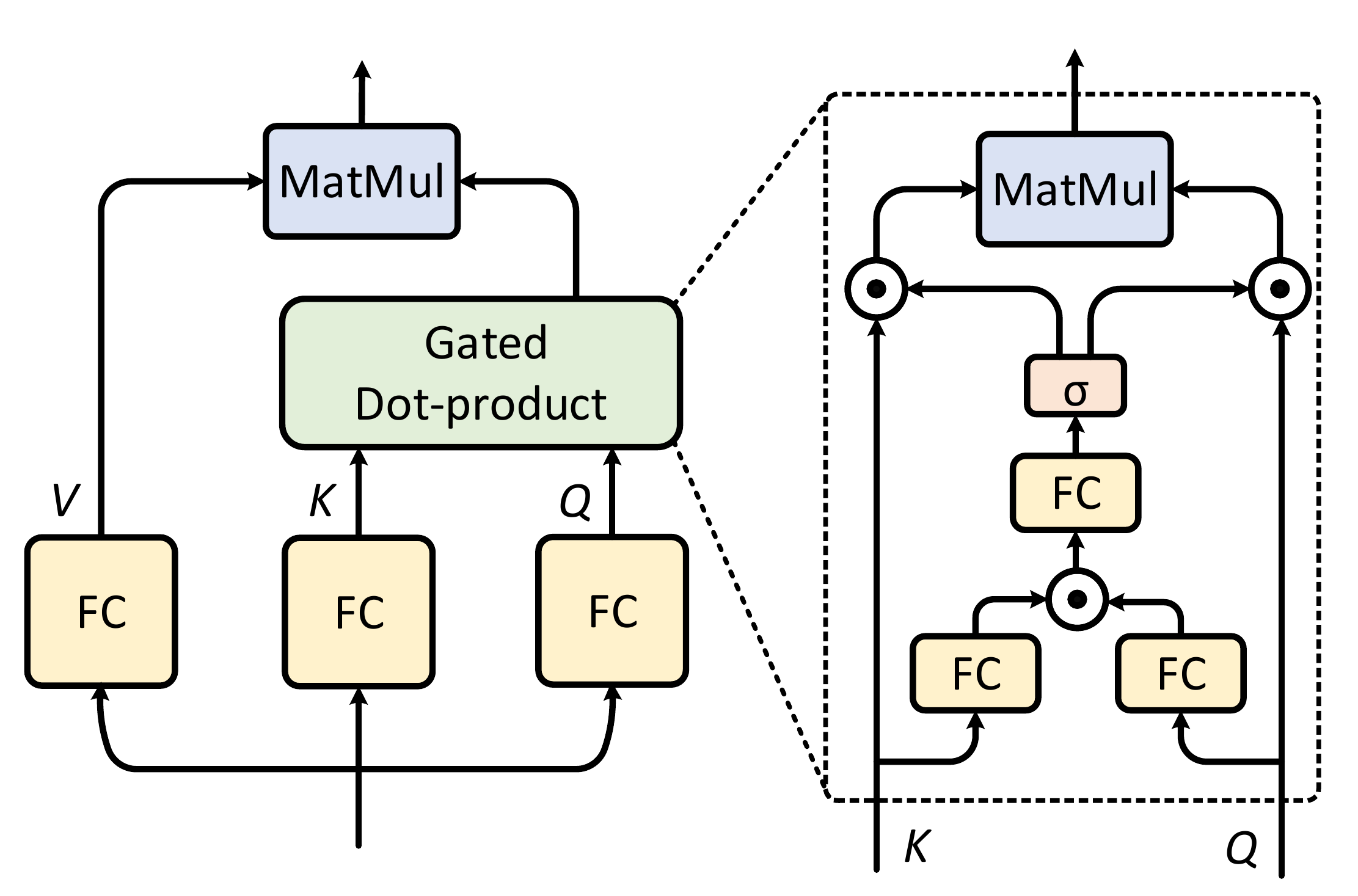}
        \caption{Gated Self-Attention (GSA)}\label{fig:gsa}
    \end{subfigure}
    \begin{subfigure}[h]{0.3\columnwidth}
        \includegraphics[width=\linewidth]{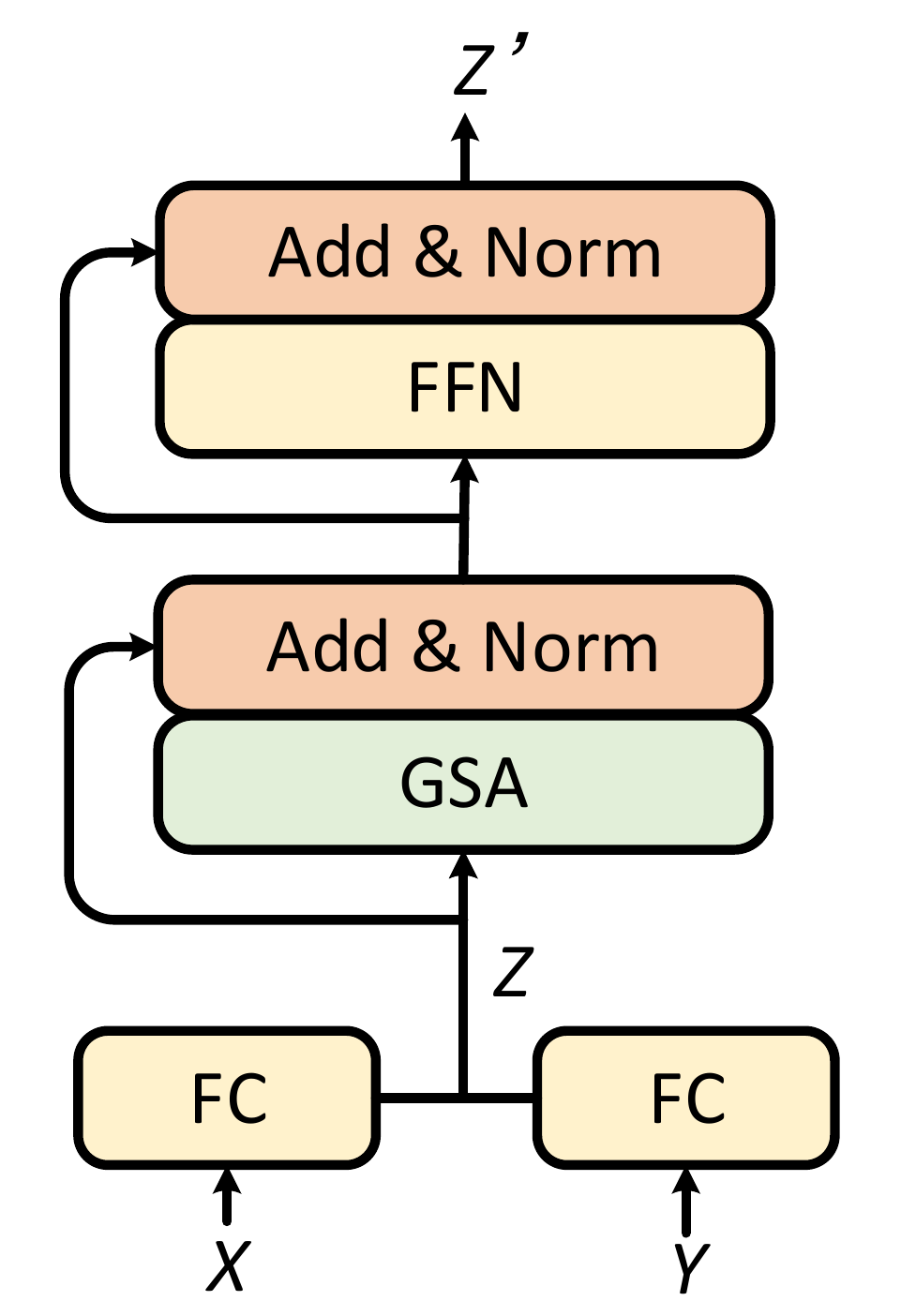}
        \caption{UA Block}\label{fig:ua_block}
    \end{subfigure}
    \caption{Flowcharts of the Gated Self-Attention (GSA) model and unified attention (UA) block for multimodal data}
    \label{fig:deep_arch}
\end{figure}

Fig. \ref{fig:gsa} illustrates the flowchart of our gated self-attention model. Similar to \cite{vaswani2017attention}, the multi-head strategy is introduced in our model to attain more diverse attention.

\subsection{Unified Attention Block}
Based on the gated self-attention model above, we introduce the multimodal unified attention block, which simultaneously models intra- and inter-modal interactions.

Given a group of textual features (\emph{e.g.}, question words) $X\in\mathbb{R}^{m\times{d_x}}$ and a group of visual features (\emph{e.g.}, image regions) $Y\in\mathbb{R}^{n\times d_y}$, we first learn two fully-connected layers $\mathrm{FC}_x$ and $\mathrm{FC}_y$ to embed $X$ and $Y$ into a $d_z$-dimensional common space, and then concatenate the two groups of embedded features on rows to form a unified feature matrix $Z$:
\begin{equation}\label{eq:unified concat}
Z = [\mathrm{FC}_x(X); \mathrm{FC}_y(Y)]
\end{equation}
where $Z=[z_1,...,z_s]\in\mathbb{R}^{s \times d_z}$ with $s=m+n$\footnote{In our implementation, we let $d_x=d_z=d$ and omit $\mathrm{FC}_x(\cdot)$ for simplicity, and rewrite Eq.(\ref{eq:unified concat}) as $Z=[X;\mathrm{FC}_y(Y)]$}.

\begin{figure*}
\begin{center}
\includegraphics[width=0.9\textwidth]{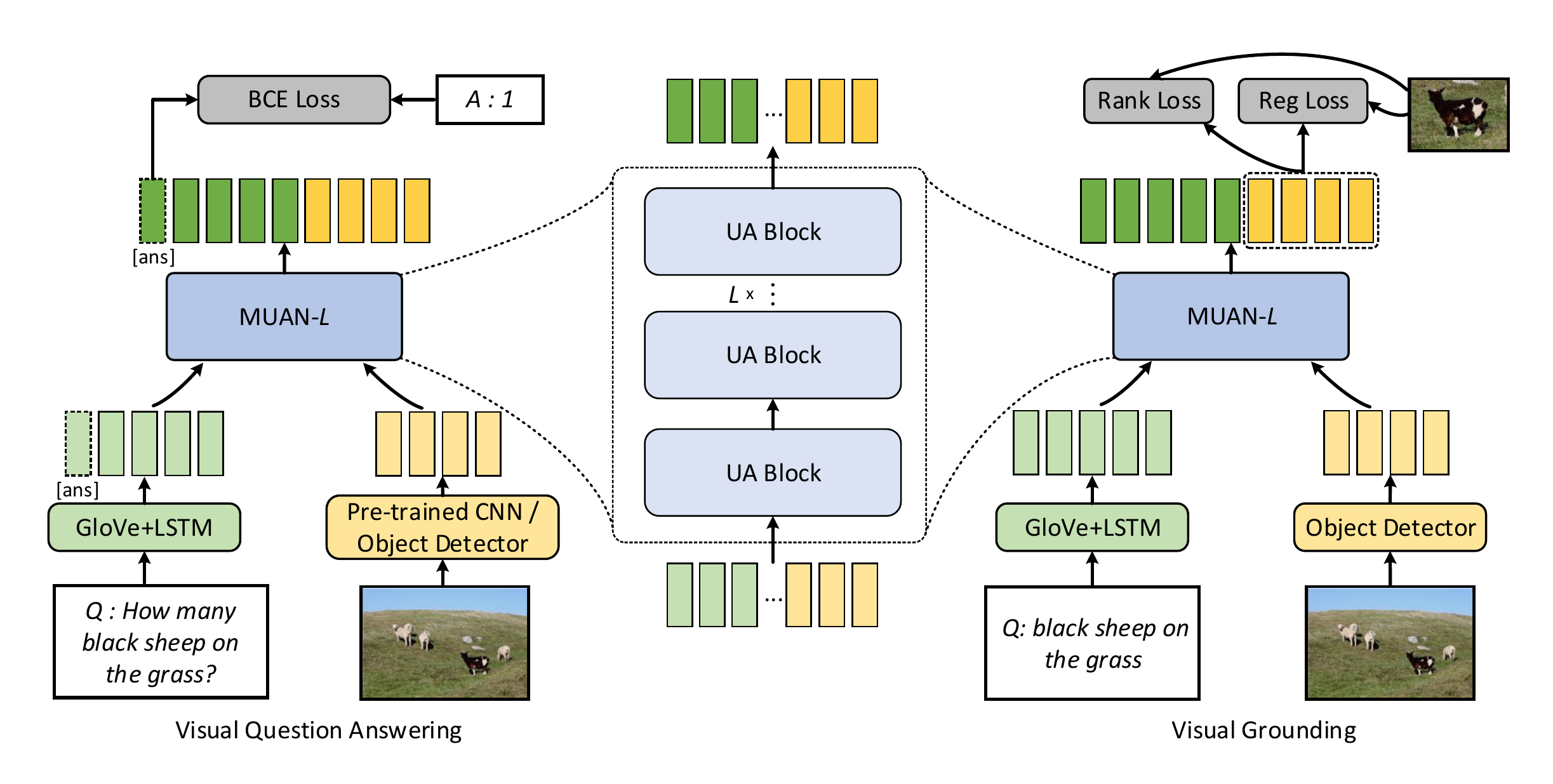}
\caption{Architectures of the Multimodal Unified Attention Networks (MUAN) for visual question answer (left) and visual grounding (right), respectively. Both architectures contain the a MUAN-$L$ model which consists of $L$ stacked UA blocks to output the features with unified attention learning. For VQA, we add a dummy token $\mathsf{[ans]}$ at the beginning of the question, and use its attended feature to predict the answer. For visual grounding, the attended features of the region proposals are used to predict their ranking scores and refined bounding boxes.}
\label{fig:muan}
\end{center}
\end{figure*}

The UA block (see Fig. \ref{fig:ua_block}) consists of a gated self-attention (GSA) module and a feed-forward network (FFN) module. Taking the unified feature matrix $Z$ as input, the GSA module learns the pairwise interactions between the sample pairs $<z_i, z_j>$ within $Z$. Since $z_i$ and $z_j$ may come from different (or the same) modalities, the intra- and inter-modal relationships are represented at the same time. Compared to existing co-attention models, which only model the inter-modal relationships \cite{kim2018bilinear}\cite{nguyen2018improved}, the intra-modal relationships (\emph{e.g.}, word-to-word or region-to-region) are also important for understanding the intrinsic structure within each modality, thus facilitating more accurate visual reasoning. The FFN module takes the output features of the GSA module as input, and then performs transformation through two consecutive fully-connected layers (FC(4$d$)-ReLU-Drop(0.1)-FC($d$)). To simplify optimization, shortcut connection \cite{he2015deep} and layer normalization \cite{ba2016layer} are applied after the GSA and FFN modules. It is worth noting that the final output features $Z'$ of the UA block are of the same shape as the input features $Z$, making it possible to stack multiple UA blocks in depth\footnote{For multiple UA blocks stacked in depth, only the first block needs to handle multimodal inputs. Eq.(\ref{eq:unified concat}) is omitted in the other blocks.}.

\section{Multimodal Unified Attention Networks}
In this section, we describe the MUAN architectures for VQA and visual grounding (see Fig. \ref{fig:muan}). The core component of both models is the deep MUAN-$L$ model, which consists of $L$ UA blocks stacked in depth to perform deep multimodal reasoning and attentional feature transformation. The proposed VQA model and the visual grounding model are very similar to each other, except for the input feature representations and the loss functions used during model training. We therefore highlight these two parts in each model.

\subsection{Architecture for VQA}
\noindent \textbf{Image and Question Representations.} The inputs for VQA consist of an images and a question, and the goal is to predict an answer to the question. Our model first extracts representations for the image and the question and then feeds the multimodal features into the MUAN model to output their corresponding output features with unified attention learning. Finally, one of the attended feature is fed to a multi-label classifier to predict the correct answer.

The input question is first tokenized into a sequence of words, and then trimmed (or zero padded) to a maximum length of $m$. Similar to \cite{devlin2018bert}, we add a dummy token $[\mathsf{ans}]$ at the beginning of the question, and the attended feature of this token will be used to predict the answer. These words are firstly represented as one-hot vectors and then transformed to 300-D word embeddings using the pre-trained GloVe model \cite{pennington2014glove}. Finally, the word embeddings are fed into a one-layer LSTM network \cite{hochreiter1997long} with $d_x$ hidden units, resulting in the final question feature ${X}\in\mathbb{R}^{(m+1) \times d_x}$. The input image is represented as a group of $d_y$-dimensional visual features extracted from a pre-trained CNN model \cite{he2015deep} or a pre-trained object detector \cite{anderson2017up-down}. This results in the image feature $Y\in\mathbb{R}^{n\times d_y}$, where $n$ is the number of extracted features.

Note that we mask the zero-padded features during attention learning to make their attention weights all zero.

\noindent \textbf{MUAN-$L$.} The multimodal features $X$ and $Y$ are fed into a deep MUAN-$L$ model consisting of $L$ UA blocks $[\mathrm{UA}^{(1)},\mathrm{UA}^{(2)},...,\mathrm{UA}^{(L)}]$. For $\mathrm{UA}^{(1)}$, $X$ and $Y$ are integrated by Eq.(\ref{eq:unified concat}) to obtain the initialized unified features $Z^{(0)}$, which are further fed to the remaining UA blocks in a recursive manner.
\begin{equation}\label{eq:MUAN}
Z^{(l+1)} = \mathrm{UA}^{(l+1)}(Z^{(l)})
\end{equation}
where $l\in[0,L-1]$. Note that the final output features $Z^{(L)}$ are the same shape as the input features $Z^{(0)}$, and each paired $<z^{(0)}_i,z^{(L)}_i>$ has a one-to-one correspondence.
\\
\textbf{Answer Prediction.} Using the attended features $Z^{(L)}$ from MUAN-$L$, we project the first feature $z^{(L)}_1$ (the $\mathsf{[ans]}$ token) into a vector $p\in\mathbb{R}^k$, where $k$ corresponds to the size of the answer vocabulary.

For the datasets that have multiple answers to each question, we following the strategy in \cite{teney2017tips} and use the binary cross-entropy (BCE) loss to train an $k$-way classifier with respect to the ground-truth label $y\in\mathbb{R}^k$:
\begin{equation}\label{eq:bce}
\mathcal{L}= \sum\limits_{i=1}^k\left(y_i\mathrm{log}(\sigma(p_i)) + (1-y_i)\mathrm{log}(1-\sigma(p_i))\right)
\end{equation}
where $\sigma(\cdot)$ is the sigmoid activation function.

For the datasets that have exactly one answer to each question, we use the softmax cross-entropy loss to train the model with respect to the one-hot ground-truth label $y\in\{0,1\}^k$:
\begin{equation}\label{eq:softmaxce}
\mathcal{L}= -y^T\mathrm{log~softmax}(p)
\end{equation}
\subsection{Architecture for Visual Grounding}
The inputs for visual grounding consist of an image and a query. Similar to the VQA architecture above, we extract the query features ${X}\in\mathbb{R}^{m \times d_x}$ using GloVe embeddings followed by a LSTM network, and extract the region-based proposal features $Y\in\mathbb{R}^{n\times d_y}$ for the image using an pre-trained object detector. Note that we do not use the dummy token for visual grounding which is specially designed for VQA.

The multimodal input features are integrated and transformed by MUAN-$L$ to output their attended representations. On top of the attended feature for each region proposal, we append two fully-connected layers to project each attended feature $z^{(L)}\in Z^{(L)}$ into a score $s\in\mathbb{R}$ and a 4-D vector $t\in\mathbb{R}^4$ to regress the refined bounding box coordinates for the proposal, respectively.
\begin{equation}\label{eq:vg}
s = \mathrm{FC}(z^{(L)}) ;~~ t = \mathrm{FC}(z^{(L)})
\end{equation}

Accordingly, a ranking loss $L_\mathrm{rank}$ and a regression loss $L_\mathrm{reg}$ are designed to optimize the model in a multitask learning manner. Following the strategy in \cite{yu2018rethinking}, KL-divergence is used as the ranking loss:
\begin{equation}\label{eq:kld_loss}
\mathcal{L}_{\textrm{rank}}= \frac{1}{n}\sum\limits_{i=1}^ns_i^*\mathrm{log}(\frac{s_i^*}{s_i})
\end{equation}
where $S=[s_1, s_2,...,s_n]\in\mathbb{R}^n$ are the predicted scores for $n$ proposals. The ground-truth label $S^*=[s^*_1, s^*_2,...,s^*_n]\in\mathbb{R}^n$ is obtained by calculating the IoU scores of all proposals w.r.t. the unique ground-truth bounding box and assign the IoU score of the $i$-th proposal to $s_i^*$ if the IoU score is larger than a threshold $\eta$ and 0 otherwise. Softmax normalizations are respectively applied to $S$ and $S^*$ to make them form a score distribution.

The smoothed $\ell_1$ loss \cite{girshick2015fast} is used as the regression loss to penalize the differences between the refined bounding box and the ground-truth bounding box:
\begin{equation}\label{eq:reg_loss}
\mathcal{L}_{\textrm{reg}}= \frac{1}{n}\sum\limits_{i=1}^n \mathrm{smooth}_{L_1}(t_i^*, t_i)
\end{equation}
where  $t_i\in\mathbb{R}^4$ and $t_i^*\in\mathbb{R}^4$ correspond to the coordinates of the predicted bounding box and the ground-truth bounding box for $i$-th proposal, respectively.

By combining the two terms, we obtain the overall loss function $L_\mathrm{all}$ as follows:
\begin{equation}\label{eq:VGD loss}
\mathcal{L}_\mathrm{all} = \mathcal{L}_\mathrm{rank} + \lambda \mathcal{L}_\mathrm{reg}
\end{equation}
where $\lambda$ is a hyper-parameter to balance the two terms.
\section{Experiments}
In this section, we conduct experiments to evaluate the performance of the MUAN models in VQA and visual grounding tasks. We conduct extensive ablation experiments to explore the effect of different hyper-parameters in MUAN. Finally, we compare the best MUAN models to current state-of-the-art methods on five benchmark datasets (two VQA datasets and three visual grounding datasets).

\captionsetup[subfigure]{font=small}
\begin{figure*}
    \centering
    \begin{subfigure}[h]{0.49\textwidth}
        \includegraphics[width=\linewidth]{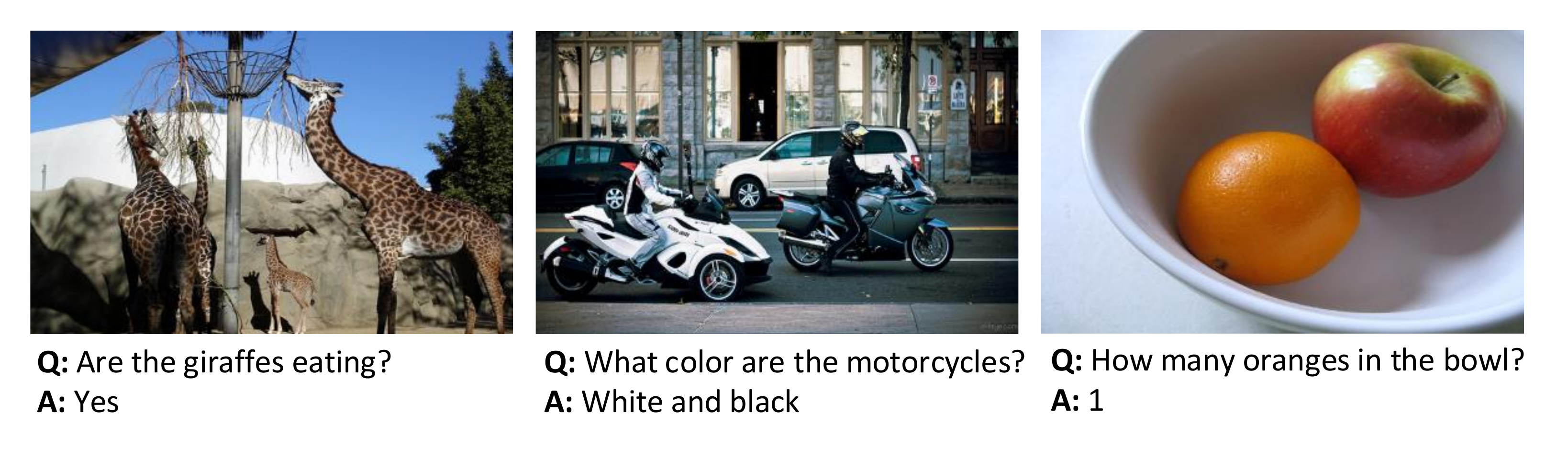}
        \vspace{-20pt}
        \caption{VQA-v2}
    \end{subfigure}
    \begin{subfigure}[h]{0.49\textwidth}
        \includegraphics[width=\linewidth]{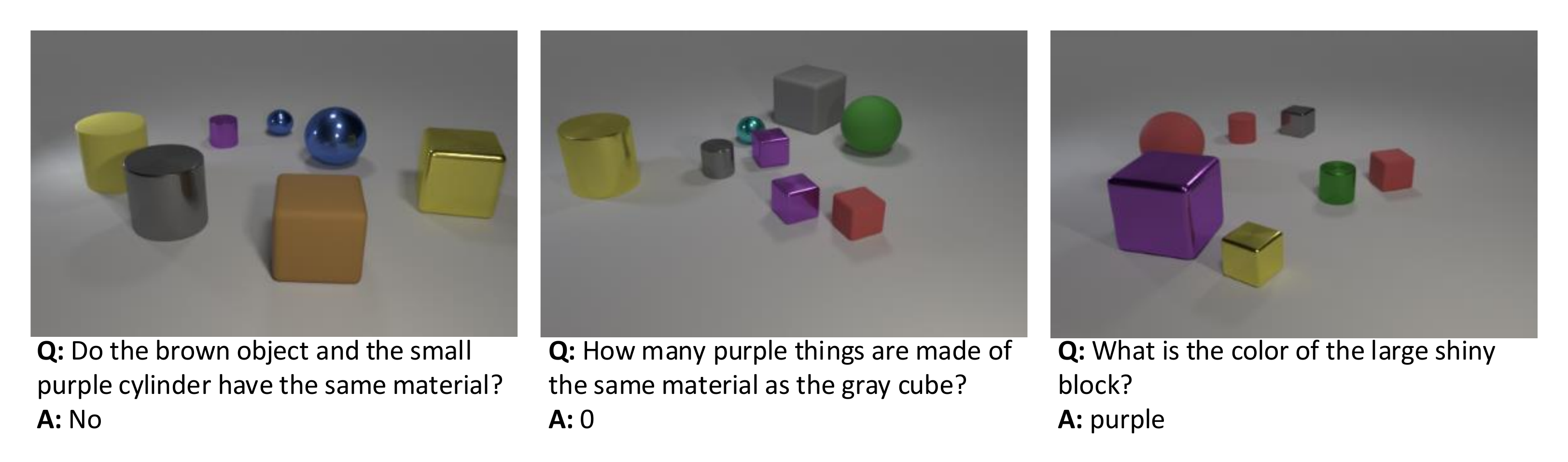}
        \vspace{-20pt}
        \caption{CLEVR}
    \end{subfigure}
    \begin{subfigure}[h]{0.32\textwidth}
        \includegraphics[width=\linewidth]{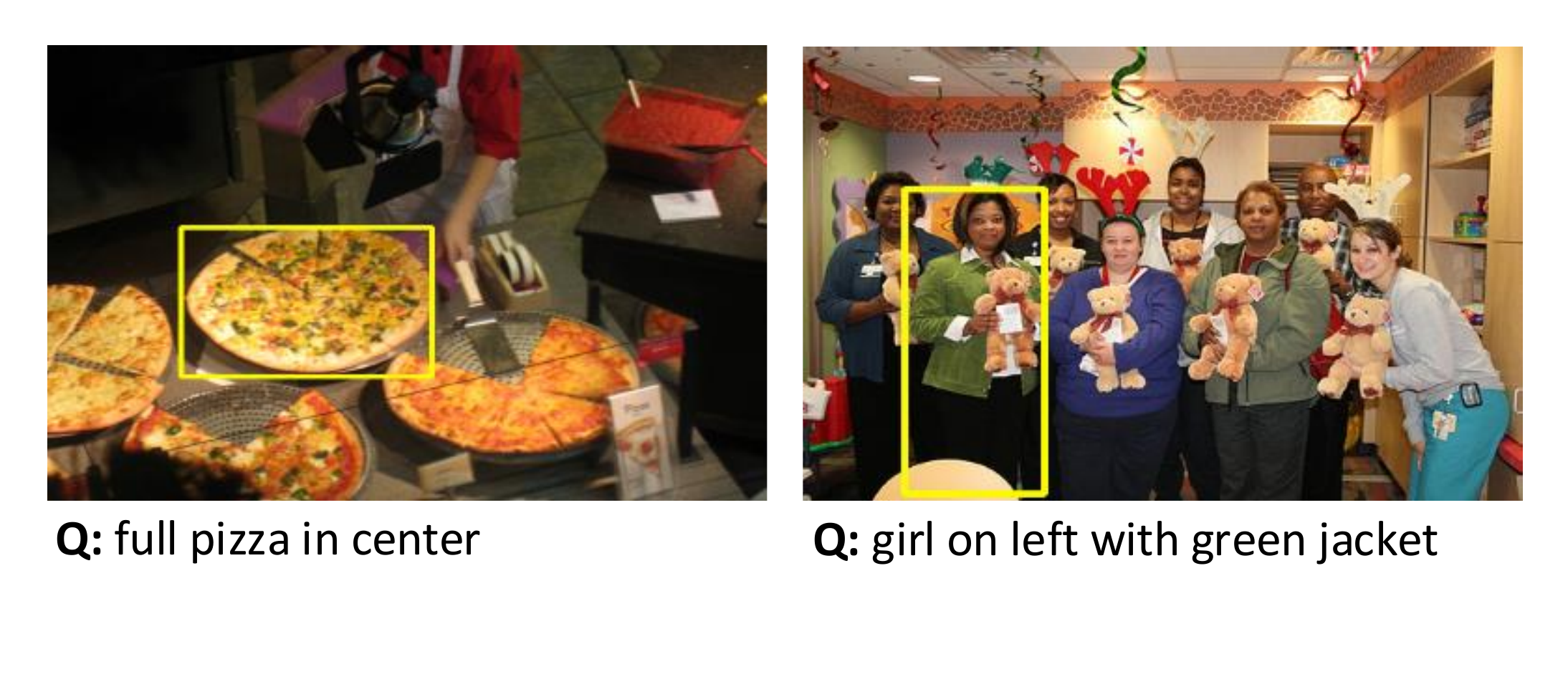}
        \vspace{-20pt}
        \caption{RefCOCO}
    \end{subfigure}
    \begin{subfigure}[h]{0.32\textwidth}
        \includegraphics[width=\linewidth]{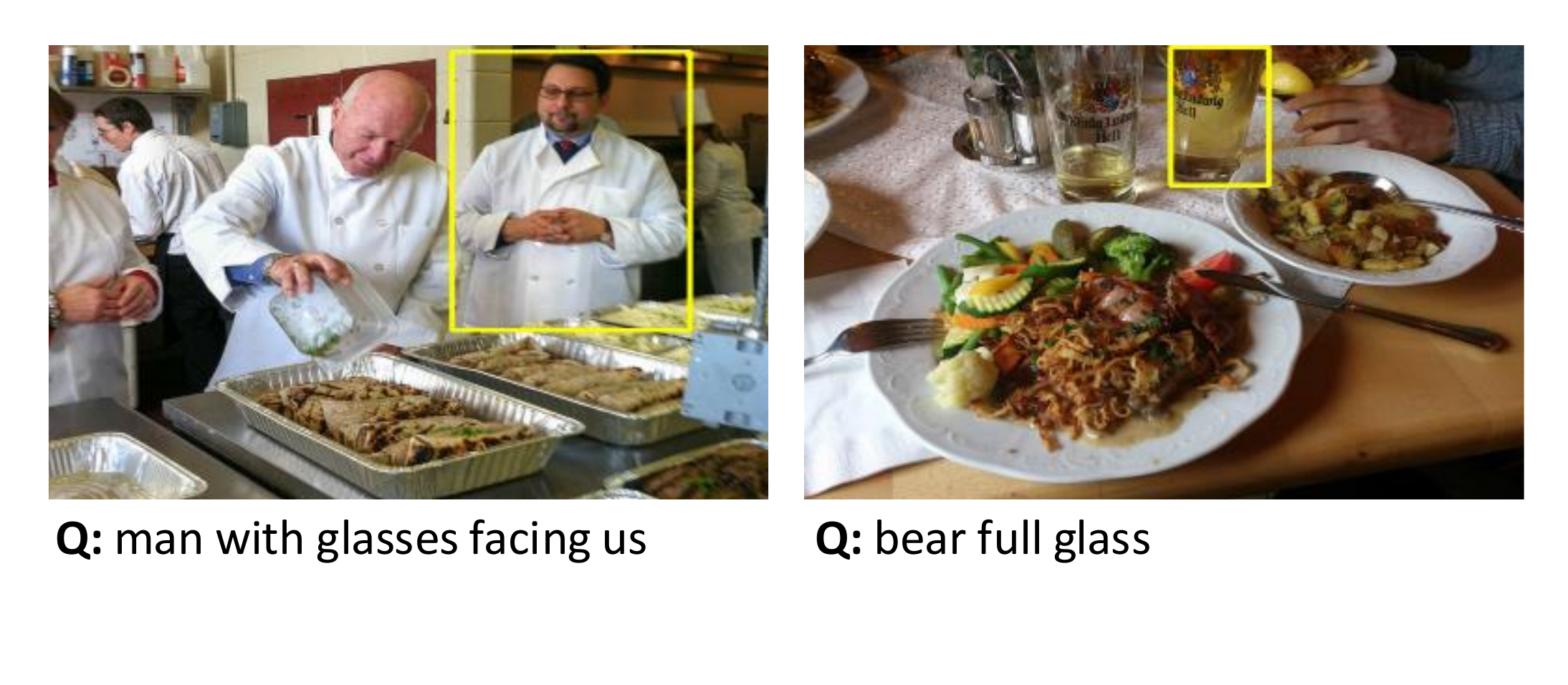}
        \vspace{-20pt}
        \caption{RefCOCO+}
    \end{subfigure}
    \begin{subfigure}[h]{0.32\textwidth}
        \includegraphics[width=\linewidth]{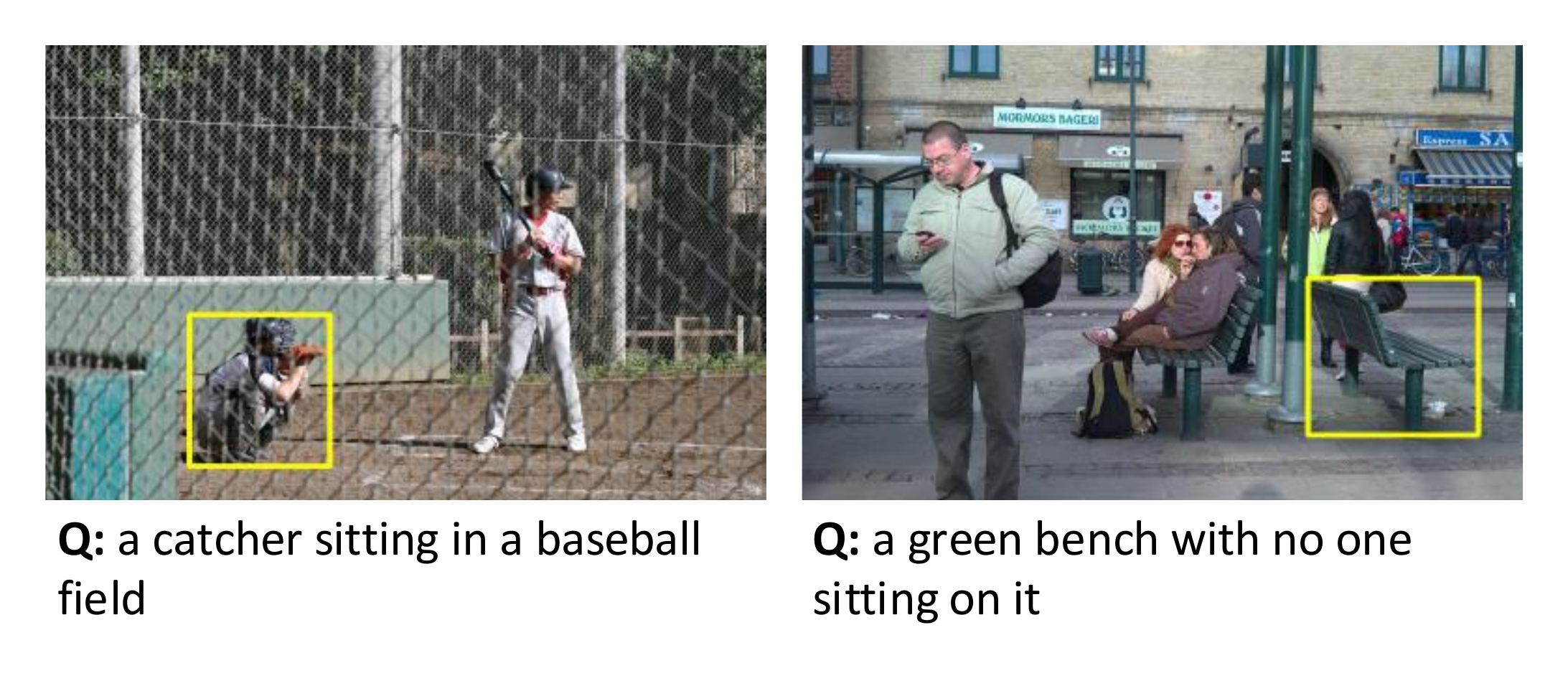}
        \vspace{-20pt}
        \caption{RefCOCOg}
    \end{subfigure}
    \caption{Typical examples from VQA-v2, CLEVR, RefCOCO, RefCOCO+, and RefCOCOg.}
    \label{fig:datasets}
\end{figure*}


\subsection{Datasets}
\noindent \textbf{\textit{VQA-v2}} is a commonly-used benchmark dataset for open-ended VQA \cite{goyal2016making}. It contains human annotated question-answer pairs for MS-COCO images \cite{lin2014microsoft}. The dataset is split into three subsets: {train} (80k images with 444k questions); {val} (40k images with 214k questions); and {test} (80k images with 448k questions). The {test} subset is further split into {test-dev} and {test-std} sets that are evaluated online with limited attempts. For each questions, multiple answer are provided by different annotators. To evaluate the performance of a model with respect to such multi-label answers, an accuracy-based evaluation metric is defined as follows which is robust to inter-human variability in phrasing the answer $a$ \cite{antol2015vqa}:
\begin{equation}\label{eq:vqa-eval}
\mathrm{Accuracy}(a) = \mathrm{min} \left\{ \frac{\mathrm{count}(a)}{3},1 \right\}
\end{equation}
where $\mathrm{count}(a)$ is a function that count the answer $a$ voted by different annotators.

\noindent \textbf{\textit{CLEVR}} is a synthesized dataset containing 100k images and 853k questions \cite{johnson2017clevr}. Each image contains 3D-rendered objects and is associated with a number of questions that test various aspects of visual reasoning including attribute identification, object counting, and logical operations. The whole dataset is split into three subsets: train (70k images with 700k questions), val (15k images with 150k questions) and test (15k images with 15k questions). Each question is associated with exactly one answer and standard accuracy metric is used to evaluate model performance.

\noindent \textbf{\textit{RefCOCO, RefCOCO+, and RefCOCOg}} are three datasets to evaluate visual grounding performance. All three datasets are collected from MS-COCO images \cite{lin2014microsoft}, but the queries are different in three respects: 1) RefCOCO \cite{kazemzadeh2014referitgame} and RefCOCO+ \cite{kazemzadeh2014referitgame} contains short queries (3.6 words on average) while RefCOCOg \cite{mao2016generation} contains relatively long queries (8.4 words on average); 2) RefCOCO and RefCOCO+ contain 3.9 same-type objects on average, while in RefCOCOg this number is 1.6; and 3) RefCOCO+ does not contain any location word, while the counterparts do not have this constraint.
RefCOCO and RefCOCO+ are split into four subsets: train (120k queries), val (11k queries), testA (6k queries about people), and testB (5k queries about objects). RefCOCOg is split into three subsets: train (81k queries), val (5k queries), and test (10k queries). For all the three datasets, accuracy is adopted as the evaluation metric, which is defined as the percentage in which the predicted bounding box overlaps with the ground-truth bounding box by IoU$>$0.5.

Fig. \ref{fig:datasets} shows some typical examples from these datasets.

\subsection{Experimental Setup}
\noindent \textbf{Universal Setup.} We use the following hyper-parameters as the default settings for MUAN unless otherwise noted. In each UA block, the latent dimensionality $d$ is 768 and the number of heads $h$ is 8, so the dimensionality of each head is $d/h=96$. The latent dimensionality in the gating model $d_g$ is 96. The number of UA blocks $L$ ranges from 2 to 12.

All the models are optimized using the Adam solver \cite{kingma2014adam} with $\beta_1=0.9$ and $\beta_2=0.99$. The models (except those for CLEVR) are trained up to 13 epochs with a batch size 64 and a base learning rate $\alpha$ set to $1.5e^{-2}/\sqrt{dL}$.  Similar to \cite{kim2018bilinear}, the learning rate is warmed-up for 3 epochs and decays by 1/5 every 2 epochs after 10 epochs. We report the best results evaluated on the validation set. For CLEVR, a smaller base learning rate $\alpha =3.5e^{-3}/\sqrt{dL}$ is used to train up to 20 epochs and decay by 1/5 at the 16th and 18th epochs, respectively.

\noindent \textbf{VQA Setup.} For VQA-v2, we follow the strategy in \cite{teney2017tips} and extract the \texttt{pool5} feature for each object from a Faster R-CNN model (with a ResNet-101 backbone) \cite{ren2015faster} pre-trained on the Visual Genome dataset \cite{krishna2016visual}, resulting in the input visual features $Y\in\mathbb{R}^{n\times 2048}$, where $n\in[10,100]$ is the number of extracted objects with a confidence threshold. The maximum number of question words $m=14$, and the size of the answer vocabulary $k=3129$, which corresponds to answers appearing more than 8 times in the training set. For CLEVR, we follow the strategy in \cite{hudson2018compositional} and extract the \texttt{res4b22} features from a ResNet-101 model pre-trained on ImageNet \cite{he2015deep}, resulting in the image features $Y\in\mathbb{R}^{196\times 1024}$. The maximum number of question words $m=43$, and the size of the answer vocabulary $k=28$.

\noindent \textbf{Visual Grounding Setup.} We use the same settings for the three evaluated datasets. To detect proposals and extract their visual features for each image, we use two pre-trained proposal detectors as previous works did: 1) a Faster R-CNN model \cite{ren2015faster} pre-trained on the Visual Genome dataset \cite{yu2018rethinking}; and 2) a Mask R-CNN model \cite{he2017mask} pre-trained on MS-COCO dataset \cite{yu2018mattnet}. During the training data preparation for the proposal detectors, we exclude the images in the training, validation and testing
sets of RefCOCO, RefCOCO+ and RefCOCOg to avoid contamination of the used visual grounding datasets. Each of the obtained proposal visual features is further concatenated with a spatial feature containing the bounding-box coordinates of the proposal\footnote{For each proposal, we first extract a 5-D spatial feature $[x_{tl}/W, y_{tl}/H, x_{br}/W, y_{br}/H,wh/WH]$ proposed in \cite{yu2016modeling}, and then linearly transform it to a 2048-D feature with a fully-connected layer to match the dimensionality of a 2048-D proposal visual feature.}. This results in the image features $Y\in\mathbb{R}^{100\times 4096}$. The maximum number of question words $m$ is 15 and the loss weight $\lambda$ is 0.5.

\captionsetup[subfigure]{font=small}
\begin{figure}
    \centering
    \begin{subfigure}[h]{0.49\columnwidth}
        \includegraphics[width=\linewidth]{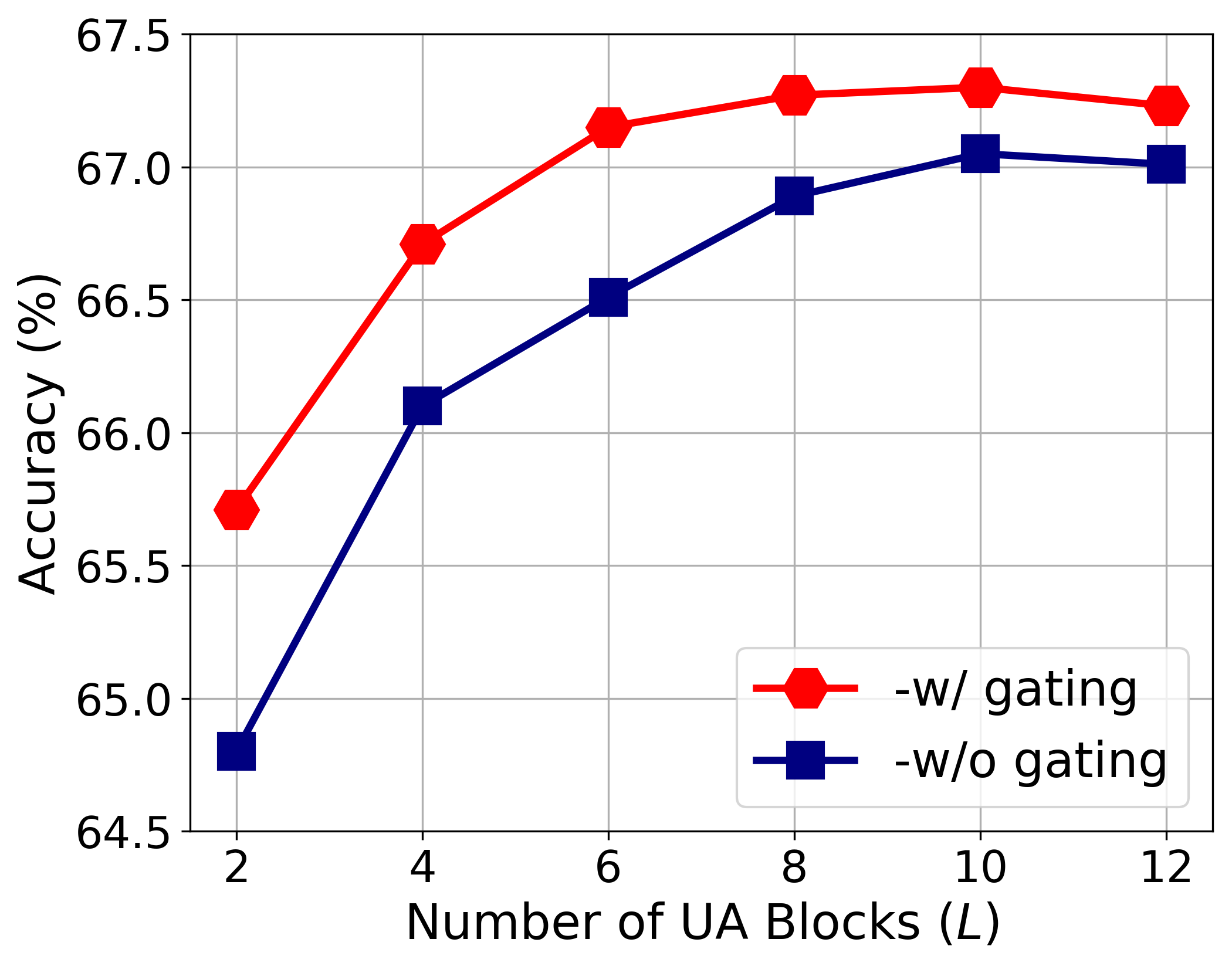}
        \caption{Effect of gating mechanism}\label{fig:aba_gate}
    \end{subfigure}
    \begin{subfigure}[h]{0.49\columnwidth}
        \includegraphics[width=\linewidth]{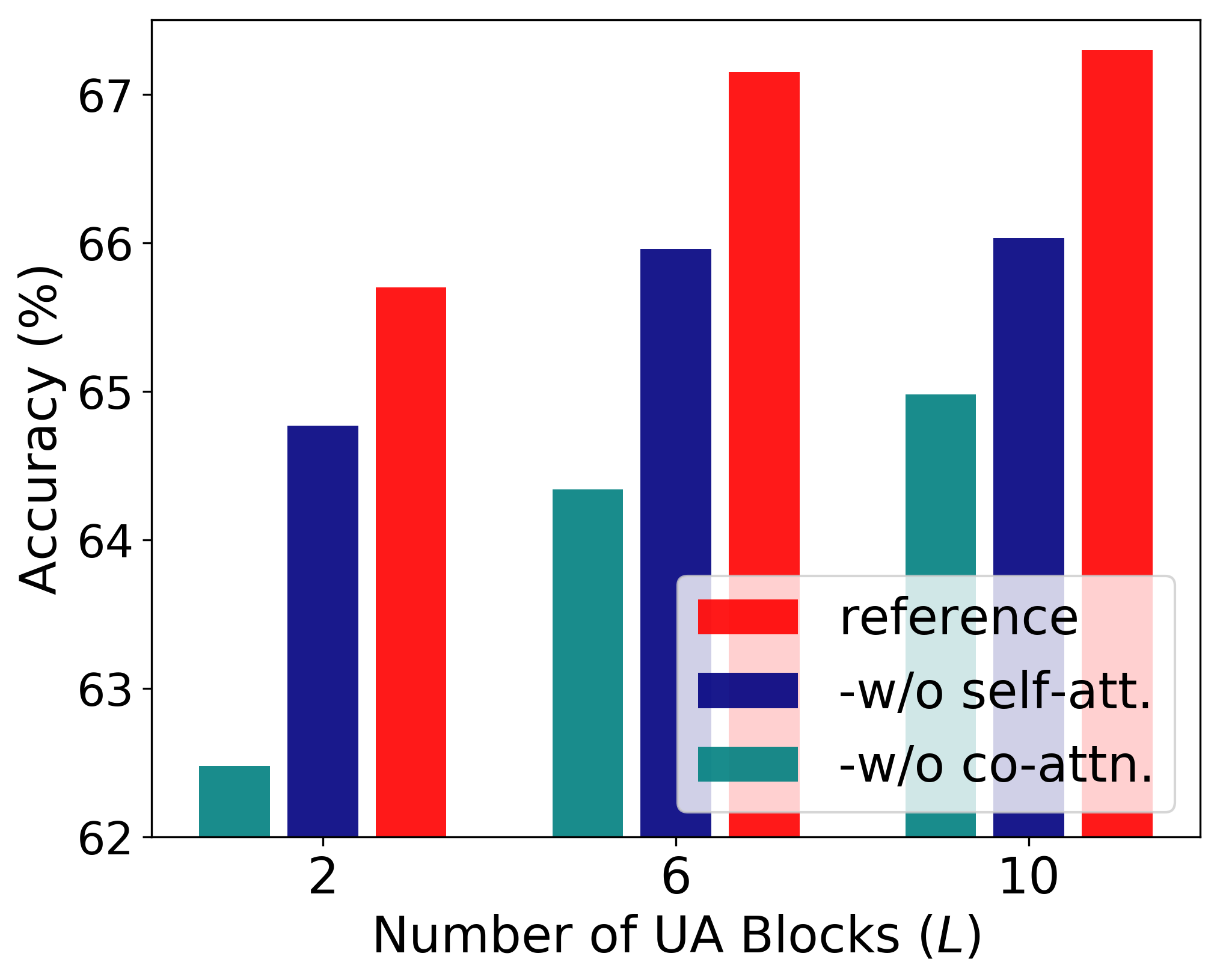}
        \caption{Effect of self- and co-attention}\label{fig:aba_saca}
    \end{subfigure}
    \caption{Ablation of the MUAN models with the number of UA blocks $L$ ranges from 2 to 12. All results are evaluated on the {val} split of VQA-v2. (a) Results of MUAN-$L$ variants with or without the gating mechanism. (b) Results of reference MUAN model along with the variants without modeling self-attention ($A_{TT}$ and $A_{VV}$) or co-attention ($A_{TV}$ and $A_{VT}$).}
    \label{fig:depth_acc}
\end{figure}

\subsection{Ablation Studies}
We run a number of ablation experiments on VQA-v2 to explore the effectiveness of MUAN.

First, we explore the effectiveness of the gating mechanism for the UA block with respect to different number of block $L$. In Fig. \ref{fig:aba_gate}, we report the overall accuracies of the MUAN-$L$ models ($L$ ranges from 2 to 12) with the gating mechanism (\emph{i.e.}, Eq.(\ref{eq:gdp})) or without the gating mechanism (\emph{i.e.}, Eq.(\ref{eq:sdp})) for the UA block. From the results, we can see that MUAN with the gating model steadily outperforms counterpart without the gating model. Furthermore, increasing $L$ consistently improves the accuracies of both models, which finally saturate at $L=10$. We think the saturation is caused by over-fitting. To train a deeper model we may require more training data \cite{devlin2018bert}.

Next, we conduct the ablation studies to explore the effects of self-attention and co-attention in MUAN. By masking the values in the self-attention part (\emph{i.e.}, $A_{TT}$ and $A_{VV}$) or the co-attention part (\emph{i.e.}, $A_{TT}$ and $A_{VV}$) to $-\infty$, we obtain two degraded variants of MUAN. We compare the two MUAN variants to its reference model in Fig. \ref{fig:aba_saca} with $L\in\{2,6,10\}$. The results shows that: 1) both the self-attention and co-attention in MUAN contribute to the performance of VQA; and 2) co-attention plays a more important role than self-attention in MUAN, especially when the model is relatively shallow.

\begin{table}
	\centering
    \footnotesize
	\caption{Ablation of the MUAN-10 models with different hyper-parameters. All results are reported on the val split of VQA-v2. Unlisted hyper-parameter values are identical to those of the reference model.}
    \begin{tabular}{c|cccc|c|c}
        \toprule
        & $d$ & $h$ & $d/h$ & $d_g$ & \makecell{Acc. (\%)} & \makecell{\#Param ($\times 10^6$)} \\
        \midrule
        ref. & 768 & 8 & 96 & 96& {67.28} & 83.0 \\
        \midrule
        \multirow{3}*{(A)} & &  &  & 32& 67.16 & 82.9 \\
        & &  &  & 64& 67.27 & 82.9 \\
        & & & & 128 & 67.17 & 83.1 \\
        \midrule
        \multirow{3}*{(B)}& & 6 & 128 && 67.11 & 83.1\\
        & & 12 & 64 && 67.23 & 82.9 \\
        & & 16 & 48 && 67.25 & 82.9\\
        \midrule
       \multirow{3}*{(C)} &256 &  & 32 && 66.30 &  14.5\\
        &512 &  & 64 && 66.92 & 40.6 \\
        &1024 &  & 128 & & \textbf{67.30}& 141.6 \\
        \bottomrule
    \end{tabular}
    \label{table:aba_gsa}
\end{table}

Finally, we investigate MUAN-10 model performance with different hyper-parameters for the UA block in Table \ref{table:aba_gsa}. In row (A), we vary the dimensionality $d_g$ in the gating model. The results suggest that the reference model results in a 0.12 point improvement over the worst counterpart. Further, the model sizes of these variants are almost identical, indicating that the computational cost of the gating model can be more or less ignored. In row (B), we vary the number of parallel heads $h$ with a fixed output dimensionality $d$, keeping the computational cost constant. The results suggest that $h=8$ is the best choice for MUAN. Too few or too many heads reduces the quality of learned attention. In row (C), we fix the number of heads to $h=8$ and vary the dimensionality $d$, resulting in much smaller and larger models with the model complexity proportional to $O(d^2)$. From the results, we can see that $d$ is a key hyper-parameter to the performance. Too small $d$ may restrict the model capacity, leading to inferior performance. The model with $d=1024$ slightly surpasses the reference model at the expense of much higher computational complexity and greater risk of over-fitting.

The hyper-parameters in the reference model is a trade-off between efficiency and efficacy. Therefore, we adopt the reference MUAN-10 model (abbreviated to MUAN for simplicity) in all the following experiments.

\begin{table}
\centering
\footnotesize
\caption{Accuracies (\%) of the \emph{single-model} on the {test-dev} and {test-std} splits of VQA-v2 to compare with the state-of-the-art methods. All models use the same bottom-up attention visual features \cite{anderson2017up-down} and are trained on the {train+val+vg} splits, where vg indicates the augmented training samples from Visual Genome \cite{krishna2016visual}.}
\label{table:vqav2}
\begin{tabular}{l|cccc|c}
\toprule
\multirow{3}{*}{Method}& \multicolumn{4}{c|}{Test-dev} &Test-std \\
\cmidrule{2-6}
 & All & Y/N & Num & Other & All \\
\midrule
Bottom-Up \cite{teney2017tips} &65.32& 81.82& 44.21& 56.05 &65.67\\
Counter \cite{zhang2018learning} & 68.09 &83.14 & 51.62& 58.97  &68.41 \\
MFH+CoAtt \cite{yu2018beyond}& 68.76 &84.27 &49.56 &59.89& -\\
BAN \cite{kim2018bilinear}& 69.52 &85.31& 50.93& 60.26&-\\
BAN+Counter \cite{kim2018bilinear}& 70.04& 85.42& {54.04}&60.52&70.35 \\
DFAF \cite{peng2018dynamic} & 70.22& 86.09& 53.32 &60.49 & 70.34 \\
MCAN \cite{yu2019mcan} & 70.63 & \textbf{86.82} & 53.26 & 60.72 & 70.90 \\
\midrule
MUAN (ours) & \textbf{70.82} & {86.77} &\textbf{54.40} &\textbf{60.89}  & \textbf{71.10}\\
\bottomrule
\end{tabular}
\end{table}

\begin{table*}
\centering
\footnotesize
\caption{Overall accuracies (\%) on the {test} split of CLEVR to compare with the state-of-the-art methods. (*) denotes use of extra program labels. ($\dag$) denotes use of data augmentation.}
\label{table:clevr}
\begin{tabular}{c|ccccccccc|c}
\toprule
{Method} & \makecell{Human\\\cite{johnson2017clevr}} & \makecell{Q-type Prior \\\cite{johnson2017clevr}} & \makecell{LSTM\\\cite{johnson2017clevr}}& \makecell{CNN+LSTM\\\cite{johnson2017clevr}} &\makecell{N2NMN*\\\cite{hu2017learning}} &\makecell{RN$^\dag$\\\cite{santoro2017simple}}  &\makecell{PG+EE*\\\cite{johnson2017inferring}}  & \makecell{FiLM\\\cite{perez2018film}} & \makecell{MAC\\\cite{hudson2018compositional}}  & \makecell{{MUAN}\\(ours)}\\
\midrule
Accuracy & 92.6 &41.8&46.8&52.3& 83.7 & 95.5  & 96.9& 97.7 & \textbf{98.9} & 98.7 \\
\bottomrule
\end{tabular}
\end{table*}

\begin{table*}
\centering
\footnotesize
\caption{Accuracies (\%) on RefCOCO, RefCOCO+ and RefCOCOg to compare with the state-of-the-art methods. All methods use the detected proposals rather than the ground-truth bounding-boxes. COCO \cite{lin2014microsoft} and Genome \cite{krishna2016visual} denote two datasets for training the proposal detectors. SSD \cite{liu2016ssd}, FRCN \cite{ren2015faster} and MRCN \cite{he2017mask} denote the used detection models with VGG-16 \cite{simonyan2014very} or ResNet-101 \cite{he2015deep} backbones.}\label{table:refcoco}
\begin{tabular}{l|c|c|c|ccc|ccc|cc}
\toprule
\multirow{3}{*}{Method} &\multicolumn{3}{c|}{Proposal Generator}& \multicolumn{3}{c|}{RefCOCO} & \multicolumn{3}{c}{RefCOCO+} & \multicolumn{2}{|c}{RefCOCOg}\\
\cmidrule{2-12}
&Dataset&Detector&Backbone&  TestA & TestB & Val & TestA & TestB & Val & Test & Val \\
\midrule
Attr \cite{liu2017referring} &COCO& FRCN &VGG-16 & 72.0 & 57.3 & - & 58.0 & 46.2 & - & - & -\\
CMN \cite{hu2017modeling} &COCO& FRCN &VGG-16 & 71.0 & 65.8 & - & 54.3 & 47.8 & - & - & -\\
VC \cite{zhang2018grounding} &COCO& FRCN &VGG-16 & 73.3 & 67.4 & - & 58.4 & 53.2 & - & - & -\\
\textbf{Spe.}+Lis.+Rein.+MMI \cite{yu2017joint} &COCO& SSD &VGG-16 & 73.7 & 65.0 & 69.5 & 60.7 & 48.8 & 55.7 & 59.6 & 60.2\\
Spe.+\textbf{Lis.}+Rein.+MMI \cite{yu2017joint} &COCO& SSD &VGG-16 & 73.1 & 64.9 & 69.0 & 60.0 & 49.6 & 54.9 & 59.2 & 59.3\\
DDPN \cite{yu2018rethinking} &Genome& FRCN &VGG-16 & 76.9& 67.5 & 73.4& 67.0 & 50.2 & 60.1 & - & -\\
DDPN \cite{yu2018rethinking} &Genome& FRCN &ResNet-101 & 80.1 & 72.4 & 76.8 & 70.5 & 54.1 & 64.8 & 67.0 & 66.7 \\
MAttNet \cite{yu2018mattnet} &COCO& FRCN &ResNet-101 & 80.4 & 69.3 & 76.4 & 70.3 & 56.0 & 64.9 & 67.0 & 66.7 \\
MAttNet \cite{yu2018mattnet} &COCO& MRCN &ResNet-101 & 81.1 & 70.0 & 76.7 & 71.6 & 56.0 & 65.3 & 67.3 & 66.6 \\
\midrule
{MUAN} (ours)  & COCO& MRCN& ResNet-101 & 82.8 & {78.6} &81.4  & 70.5 & 62.9 & 68.9 & 71.5 & 71.0\\
{MUAN} (ours) & Genome & FRCN &ResNet-101 & \textbf{86.5} & \textbf{78.7} & \textbf{82.8} & \textbf{79.5} & \textbf{64.3} & \textbf{73.2} & \textbf{74.3} & \textbf{74.2} \\
\bottomrule
\end{tabular}
\end{table*}

\subsection{Results on VQA-v2}

Taking the ablation studies into account, we compare our best MUAN model to the state-of-the-art methods on VQA-v2 in Table \ref{table:vqav2}. With the same bottom-up-attention visual features \cite{anderson2017up-down}, MUAN significantly outperforms current state-of-the-art methods BAN \cite{kim2018bilinear} by 1.3 points in terms of overall accuracy on the test-dev split. Furthermore, for the \emph{Num}-type questions, which verify object counting performance, BAN+Counter \cite{kim2018bilinear} reports the best result by utilizing an elaborate object counting module \cite{zhang2018learning}. In contrast, MUAN achieves slightly higher accuracy than BAN+Counter, and in doing so does not use the auxiliary bounding-box coordinates of each object \cite{zhang2018learning}. This suggests that MUAN can perform accurate object counting based on the visual features alone. As far as we know, MUAN is the first single model that achieves 71\%+ accuracy on the test-std split with the standard bottom-up-attention visual features provided by \cite{anderson2017up-down}.

\subsection{Results on CLEVR}
We also conduct experiments to compare MUAN with existing state-of-the-art approaches, and human performance on CLEVR, which is a synthesized dataset for evaluating compositional visual reasoning. Compared to VQA-v2, CLEVR requires a model not only to focus on query-specific objects, but only to reason the {relations} among the related objects, which is much more challenging. In the meantime, since the image contents are completely synthesized by the algorithm, it is possible for a model to fully understand the semantic, resulting in relatively higher performance of existing state-of-the-arts compared to those on VQA-v2.

From the results shown in Table \ref{table:clevr}, we can see that MUAN is at least comparable to the state-of-the-art, even if the model is not specifically designed for this dataset. While some prior approaches used extra supervisory program labels \cite{johnson2017inferring}\cite{hu2017learning} or augmented dataset \cite{santoro2017simple} to guide training, MUAN is able to learn to infer the correct answers directly from the image and question features.

\subsection{Results on RefCOCO, RefCOCO+, and RefCOCOg}
We report the comparative results on RefCOCO, RefCOCO+, and RefCOCOg in Table \ref{table:refcoco}. We use the common evaluation criterion accuracy, which is defined as the percentage of predicted bounding box overlaps with the groundtruth of IoU $>$ 0.5. From the results, we can see that: 1) with the standard proposal features extracted from the detector pre-trained on MSCOCO, MUAN reports a remarkable improvement over MAttNet, the state-of-the-art visual grounding model; 2) with the powerful proposal features extracted from the detector pre-trained on Visual Genome, MUAN reports $\sim$9\% improvement over a strong baseline DDPN \cite{yu2018rethinking}, which uses the same visual features. These results reveal the fact that MUAN outperforms existing state-of-the-arts steadily regardless of the used proposal features. Compared with existing approaches, MUAN additionally models the intra-modal interactions within each modality, which provide contextual information to facilitate visual grounding performance.

\begin{figure*}
\begin{center}
\includegraphics[width=0.94\linewidth]{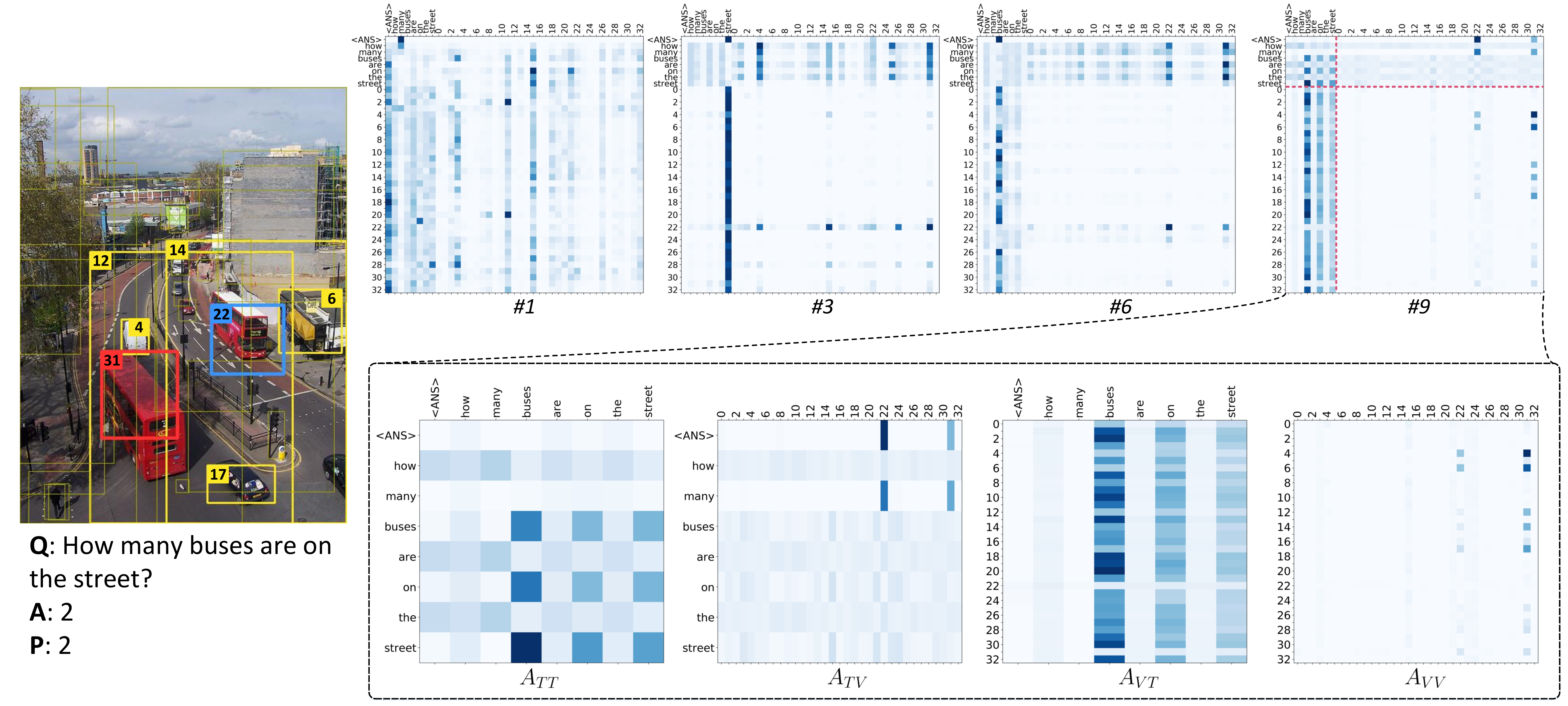}
\caption{Visualizations of the learned unified attention maps (Eq.(\ref{eq:gdp})) for VQA. The attention maps come from the 1st, 3rd, 6th and 9th UA block, respectively. The index within [0-32] on the axes of the attention maps corresponds to the object in the image (33 objects in total). For better visualization effect, we highlight the objects in the image that are related to the answer. Furthermore, we split the last attention map into four parts (\emph{i.e.}, $A_{TT}$, $A_{VT}$, $A_{TV}$ and $A_{VV}$) to carry out detailed analysis.}\label{fig:vis_vqa}
\end{center}
\end{figure*}

\begin{figure*}
\begin{center}
\includegraphics[width=0.94\linewidth]{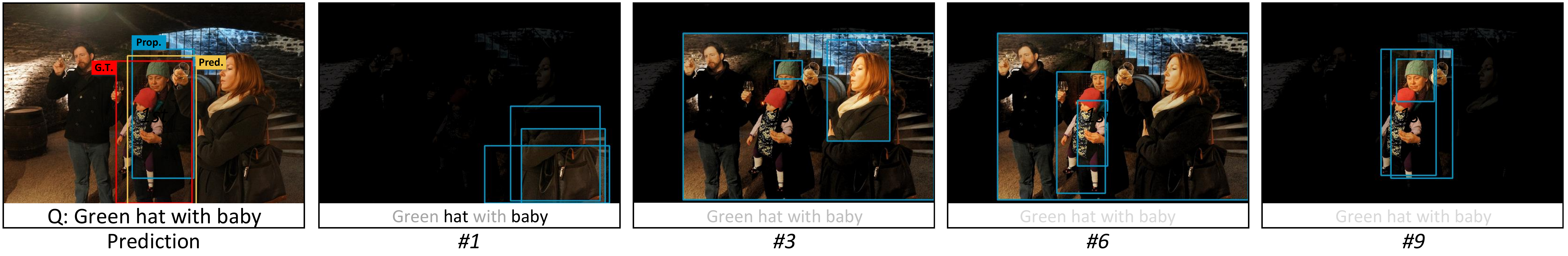}
\caption{Visualizations of the prediction and the learned visual attention for visual grounding. The groundtruth (red), top-ranked proposal (blue) and refined prediction (yellow) are shown in the first image. Next four images illustrate the learned visual attentions from the 1st, 3rd, 6th and 9th UA blocks, respectively. The visual attention is represented by three {representative} objects with the largest attention values. The brightness of objects and darkness of words represent their importance in the attention weights.}\label{fig:vis_vg}
\end{center}
\end{figure*}

\subsection{Qualitative Analysis}
In Fig. \ref{fig:vis_vqa}, we show one VQA example and visualize four attention maps (obtained by Eq.(\ref{eq:gdp})) from the 1st, 3rd, 6th and 9th UA blocks, respectively. Since only the feature of the $\mathsf{[ans]}$ token is used to predict the answer, we focus on its related attention weights (\emph{i.e.}, the first row of each attention map). In the 1st attention map, the word `many' obtains the largest weight while the other words and visual objects are almost abandoned. This suggests that the 1st block acts as a question-type classifier. In the 3rd attention map, the word `street' is highlighted, which is a contextual word to understand the question. The key word `buses' is highlighted in the 6th attention map, and the two buses (\emph{i.e.}, the 22th and 31th objects) are highlighted in the 9th attention map. This visual reasoning process explains the information of the highlighted words and objects is gradually \emph{aggregated} into the $\mathsf{[ans]}$ feature. For the 9th UA block, we split its attention map into four parts (\emph{i.e.}, $A_{TT}$, $A_{VT}$, $A_{TV}$ and $A_{VV}$). In $A_{TT}$, the largest values reflect the relationships between the key word and its context, providing a structured and fine-grained understanding of the question semantics (\emph{i.e.}, bus is on the street). In $A_{TV}$, some words on the rows attend to the key objects, suggesting that these words aggregate the information from the key objects to improve their representations. Similar observations can be observed from $A_{VV}$ and $A_{VT}$.

In Fig. \ref{fig:vis_vg}, we demonstrate one visual grounding example and visualize the prediction and the learned unified attention. In the first image, we can see that MUAN accurately localize the most relevant object proposal, and then output the refined bounding boxes as the final prediction. We visualize the learned textual and visual attentions of the 1st, 3rd, 6th and 9th UA blocks, respectively. By performing columnwise max-pooling over the unified attention map, we obtain the attention weights for the words and objects. For better visualization effect, we only visualize three representative objects with the largest attention weights. From the results, we can see that: 1) the keywords are highlighted only in the 1st block, indicating that this information has been successfully transferred to the attended visual features in the following blocks; and 2) the learned visual attention in the 1st block is meaningless. After receiving the textual information, the visual attention tends to focus on the contextual objects in the 3rd and 6th blocks (\emph{i.e.}, the hat and the baby), and finally focuses on the correct target object (\emph{i.e.}, the woman) in the 9th block.

\section{Conclusion and Future work}
In this work, we present a novel unified attention model that captures intra- and inter-modal interactions simultaneously for multimodal data. By stacking such unified attention blocks in depth, we obtain a Multimodal Unified Attention Network (MUAN), that is suitable for both VQA and visual grounding tasks. Our approach is simple and highly effective. We verify the effectiveness of MUAN on five datasets, and the experimental results show that our approach achieves top level performance on all the benchmarks without using any dataset specific model tuning.

Since MUAN is a general framework that can be applied to many multimodal learning tasks, there remains significant room for improvement, for example by introducing multitask learning with sharing the same backbone model or introducing weakly-supervised model pre-training with large-scale multimodal data in the wild.

\ifCLASSOPTIONcaptionsoff
  \newpage
\fi



\bibliographystyle{IEEEtran}
\bibliography{uan}

\end{document}